\pdfoutput=1

\documentclass[11pt]{article}

\usepackage[preprint]{acl}

\usepackage{times}
\usepackage{latexsym}

\usepackage{booktabs}
\usepackage{multirow}
\usepackage{tabularx}
\usepackage{graphicx} 
\usepackage{amsmath}  
\usepackage{hyperref} 
\usepackage{adjustbox}
\usepackage{enumitem}
\usepackage{algorithm}
\usepackage{algorithmic}
\usepackage{listings}

\usepackage{titlesec}
\titlespacing*{\subsection}{0pt}{1.0ex plus 1ex minus .2ex}{1.0ex plus .2ex}

\lstdefinestyle{json}{
    basicstyle=\small\ttfamily,
    showstringspaces=false,
    breaklines=true,
    frame=none,
    aboveskip=0pt,
    belowskip=0pt,
    xleftmargin=0pt,
    xrightmargin=0pt,
    backgroundcolor=\color{gray!10},
    numbers=left,
    numberstyle=\tiny,
    nolol=true
}
\usepackage{caption}
\usepackage{subcaption}
\usepackage{tcolorbox}
\usepackage{color}
\usepackage{colortbl}
\newcommand{\ours}{USP}
\newcommand{\backbonemodel}{Xing4.0-29B-A4B}
\newcommand{\releasedmodel}{Xing4.0-29B-Intent}
\newcommand{\releasedmodelheader}{\shortstack{Xing4.0-\\29B-Intent}}
\newcommand*{\ub}{\_\allowbreak}
\usepackage{tikz}
\usetikzlibrary{arrows.meta, positioning, shapes.geometric, calc}
\definecolor{preproc}{RGB}{219,234,254}  
\definecolor{runtime}{RGB}{255,237,213}  
\newcommand{\todofill}[1]{\textcolor{red}{[\textbf{TODO}: #1]}}
\newcommand{\suspendedfill}[1]{\textcolor{red}{[\textbf{SUSPENDED --- awaiting final \texttt{examples\_2} batch}: #1]}}
\newcommand{\tf}{\textcolor{red}{\,??\,}}

\definecolor{aliceblue}{RGB}{178, 217, 245}

\definecolor{babyblue}{RGB}{217, 239, 251}

\usepackage{pifont}

\definecolor{my_green}{RGB}{51,102,0}
\definecolor{my_red}{RGB}{204, 0, 0}

\definecolor{c3}{cmyk}{0.9081,0,0.7209,0.5255}

\newtcbox{\hlprimarytab}{on line, rounded corners, box align=base, colback=c3!10,colframe=white,size=fbox,arc=3pt, before upper=\strut, top=-2pt, bottom=-4pt, left=-2pt, right=-2pt, boxrule=0pt}
\newtcbox{\hlsecondarytab}{on line, box align=base, colback=red!10,colframe=white,size=fbox,arc=3pt, before upper=\strut, top=-2pt, bottom=-4pt, left=-2pt, right=-2pt, boxrule=0pt}

\newlength\myheight
\newlength\mydepth
\usepackage[T1]{fontenc}

\usepackage[utf8]{inputenc}

\usepackage{microtype}

\usepackage{inconsolata}

\title{Xing USP: A Cross-Family Semantic Contract for\\Information Extraction and Intent-Slot Parsing}

\author{
  Zhongqiu Li$^{1}$\thanks{These authors contributed equally.}\quad
  Boyang Wang$^{1}$\footnotemark[1]\quad
  Ruiyu Fang$^{1,2}$\footnotemark[1]\quad Shiquan Wang$^{1}$\footnotemark[1]\\
  \bfseries Di Wu$^{1}$\quad Jiayi Liang$^{1}$\quad
  Feilong Zhang$^{1}$\quad Hongfeng Du$^{1}$\\
  \bfseries Xiaomeng Huang$^{1}$\quad Mengjiao Bao$^{1}$\quad
  Luwen Pu$^{1}$\quad Xuewei He$^{1}$\\
  \bfseries Yu Zhao$^{1}$\quad Chao Wang$^{1}$\quad Kaidong Yu$^{1}$\quad Bingkai Yang$^{1}$\\
  \bfseries Qingjun Zhu$^{1}$\quad Shaoyi Du$^{2}$\quad
  Shuangyong Song$^{1}$\thanks{Corresponding authors: Shuangyong Song and Yongxiang Li.}\quad Yongxiang Li$^{1}$\footnotemark[2]\\[0.5em]
  {\normalsize\normalfont $^{1}$Xingchen AGI Lab,}\\
  {\normalsize\normalfont China Telecom Artificial Intelligence Technology (Beijing) Co., Ltd}\\
 {\normalsize\normalfont $^{2}$State Key Laboratory of Human-Machine Hybrid Augmented Intelligence, }\\
{\normalsize\normalfont Institute of Artificial Intelligence and Robotics, ~Xi'an Jiaotong University}\\
  {\small\ttfamily lizq48@chinatelecom.cn\quad wangby17@chinatelecom.cn}
}

\begin{document}
\maketitle
\begin{abstract}
Structured semantic parsing is typically judged by whether the correct category is assigned, which leaves value recovery and instance-level associations unexamined: a system may recognize an event yet bind an argument to the wrong instance, or identify an intent yet supply an incorrect slot value. We present {\ours}, a cross-family semantic contract that represents categories, fields, values, and instance associations through a common interface for named-entity recognition, relation extraction, event extraction, and intent-slot parsing. The contract is paired with five diagnostic matching views---category agreement, field recognition, field-value pairing, instance-bound assignment, and complete-instance recovery---that ask distinct questions about the same output. Using this contract, we normalize 61 datasets into a unified corpus with shared schemas, prompts, and matching rules, and compare six models on a fixed test subset. Our fine-tuned \releasedmodel{} attains $57.70$ task-balanced Field-Value Micro-F1 (equally weighted across the four task families) and ranks first in every family and on all five views; its margin over the strongest alternative widens from $2.93$ points for category matching to $20.73$ points for complete-instance matching. This widening gap shows that category agreement alone understates structural deficits. {\ours} thus offers the community a shared representation, a normalized 61-dataset corpus, and a diagnostic evaluation protocol that make structural agreement measurable and comparable across extraction and intent-slot tasks.
\end{abstract}

\section{Introduction}
Correctly identifying a category is only part of structured semantic parsing. A system may recognize an event but assign an argument to the wrong instance, or identify an intent while supplying an incorrect slot value. Such predictions can be well formed and partly correct while still failing to recover the required structure. Evaluating them calls for a representation that retains both the extracted content and the associations among its parts.

Universal information extraction (IE) provides a foundation through schema-conditioned modeling of heterogeneous extraction structures~\citep{DBLP:conf/acl/0001LDXLHSW22,ping-etal-2023-uniex,wang-etal-2024-true}, large instruction corpora~\citep{DBLP:journals/corr/abs-2402-14710}, and code-style schema definitions~\citep{knowcoder2024,gollie2024}. Unification also extends beyond conventional extraction. DeepStruct studies intent detection and dialogue state tracking alongside IE~\citep{wang-etal-2022-deepstruct}, and instruction-based SLU covers diverse spoken-language tasks~\citep{arora-etal-2024-universlu}. Hierarchical intent-slot parsing already distinguishes exact structures, substructures, and valid form~\citep{gupta-etal-2018-semantic}. The question for this work is therefore how to preserve instance associations and examine structural disagreement consistently across extraction and intent-slot outputs, rather than whether heterogeneous tasks can share a format at all.

We introduce {\ours}, a unified schema-driven semantic parsing resource built around a shared contract for corpus normalization, schema conditioning, structured output, and evaluation. The representation pairs category labels with groups of named argument values and retains relation, event, or call-instance associations supplied by the source annotations. The same output can then be examined through five matching views---category, field, field-value, instance-bound field, and complete instance---with parse success reported separately. These views ask different questions about the prediction, from category agreement to complete-instance recovery.

Using this contract, we normalize 61 public datasets and compare six models on a fixed test subset drawn from the collection. Under the failure-inclusive evaluation protocol, \releasedmodel{} obtains the highest score under each F1 view, with a wider margin for complete instances than for categories. This variation motivates examining which aspects of structure a model recovers, rather than relying on one aggregate score. Experimental settings specify how unsuccessful requests enter the comparison.

Our contributions are threefold:
\begin{itemize}[leftmargin=*,itemsep=2pt]
\item \textbf{An instance-preserving cross-family contract.} USP gives IE and intent-slot outputs a shared category--field--value representation that retains instance associations.
\item \textbf{A normalized corpus and diagnostic interface.} The inventory spans 61 public datasets. Five matching views and a separate parse-success measure support a consistent examination of categories, fields, values, binding, and complete instances across task families.
\item \textbf{An empirical analysis of structural disagreement.} The six-model comparison shows that category-level agreement, value matching, complete-instance recovery, and parse success distinguish different aspects of prediction quality.
\end{itemize}

\section{Related Work}
\label{sec:related}

\subsection{Universal Information Extraction}

Universal IE shares representations and learning mechanisms across extraction tasks. UIE introduces schema-conditioned text-to-structure generation~\citep{DBLP:conf/acl/0001LDXLHSW22}; UniEX models heterogeneous targets through span detection, classification, and association~\citep{ping-etal-2023-uniex}; and TRUE-UIE reduces multiple extraction tasks to two universal relations~\citep{wang-etal-2024-true}. IEPile develops large-scale schema-conditioned instruction data~\citep{DBLP:journals/corr/abs-2402-14710}, while KnowCoder represents taxonomies, definitions, and constraints through Python classes~\citep{knowcoder2024}. UC-UIE combines a unified frame-and-slots representation with hierarchical experts for judging, locating, and associating~\citep{zhou-etal-2026-capability}. MR-UIE combines a unified schema representation with multi-perspective reasoning and reinforcement learning for extraction~\citep{li2025mruie}. Together, these studies establish shared structure, schema conditioning, and reusable extraction operations as foundations for this work. Task-specific research also addresses forgetting in continual relation extraction through classifier expansion, dual knowledge distillation, and dynamic weighting~\citep{mao2025classifier}.

The task boundary is not exclusive to traditional IE. DeepStruct evaluates extraction, intent detection, and dialogue state tracking through a shared structure-prediction model~\citep{wang-etal-2022-deepstruct}. GoLLIE's class-based representation also accommodates slot filling, which is included in its training even though the main evaluations focus on entity and event tasks~\citep{gollie2024}. USP builds on these precedents by studying a common category--field--value--instance contract and applying the same named diagnostic views across the four families evaluated here. Its operational requirements describe the output contract rather than learned capability experts or a newly identified reasoning decomposition.

IEPile is the closest corpus comparison. Its schema-conditioned construction and task-specific evaluation provide a direct reference for organizing heterogeneous extraction data. Our inventory overlaps substantially with its training and held-out sources. The comparison therefore concerns the common contract and diagnostic coverage used in USP, rather than inventory size alone. In particular, IEPile's event examples already group triggers and arguments by instance; USP's use of instance associations should not be read as their introduction to IE. Different output representations and matching protocols also preclude a direct comparison of published scores.

\subsection{Schema Semantics and Robustness}

Schema definitions communicate more than label names. GoLLIE encodes annotation guidelines~\citep{gollie2024}, and KnowCoder includes descriptions and constraints in code-style schemas~\citep{knowcoder2024}. SGD-X tests paraphrases of schema names and descriptions~\citep{sgdx2022}; Description Boosting examines description wording in zero-shot entity and relation extraction~\citep{picco-etal-2024-description}; and RUIE-Bench evaluates perturbations across universal IE tasks~\citep{zhu-etal-2025-towards}. These studies motivate a distinction between the presence of descriptions and evidence that they improve parsing. USP documents the descriptions in its retained prompts, while controlled description sensitivity and schema perturbation remain outside the completed experiments.

Schema retrieval offers a complementary approach to large-schema inputs~\citep{liang-etal-2025-schema,ho-etal-2026-schemarag}. The present study keeps candidate reduction outside its contribution and evaluation scope.

\subsection{Spoken Language Understanding and Structured Evaluation}

Intent classification has long supported user-query understanding in service systems~\citep{song2017intension}. Intent-slot parsing has its own history of shared infrastructure and compositional representations. OpenSLU provides reusable SLU infrastructure~\citep{qin-etal-2023-openslu}, and UniverSLU uses natural-language instructions across diverse spoken-language tasks~\citep{arora-etal-2024-universlu}. TOP represents compositional intents and slots hierarchically and evaluates exact trees, brackets, subtrees, and tree validity~\citep{gupta-etal-2018-semantic}. ATIS~\citep{hemphill1990atis}, SNIPS~\citep{coucke2018snips}, MASSIVE~\citep{massive2022}, and MultiWOZ~\citep{multiwoz2018} provide the intent, slot, multilingual, and dialogue-state annotations used in this broader setting.

INT models cross-language information transfer for multilingual intent detection and slot filling~\citep{wu2025int}, while a recent survey reviews multi-intent SLU methods and their intent--slot interactions~\citep{wu2025multiintent}. Adjacent work studies few-shot joint empathy detection and empathy intent classification~\citep{jiang2025utterance}.

Related hierarchical text classification methods organize category labels in a taxonomy. ToSA uses tree-aware sequence generation~\citep{zhao2023tosa}, UMP-MG uses directional message passing~\citep{ning2023umpmg}, and Dual Prompt Tuning applies layer-wise contrastive learning~\citep{xiong2024dual}. These label hierarchies differ from the category--argument instance associations examined here.

Structured evaluation must specify which aspects of a representation are compared. TOP's structural metrics already distinguish partial agreement from complete structure~\citep{gupta-etal-2018-semantic}; AMR metric research further examines how alignment and matching choices affect semantic comparisons~\citep{opitz-etal-2020-amr}. USP adopts this concern at a cross-family scope. Its diagnostic profile examines categories, fields, values, binding, and complete instances from the same outputs, with parse success, per-dataset results, and missingness reported separately. The contribution is this concrete shared contract and diagnostic application, not the general idea that valid form and correct content differ.

\begin{figure*}[!t]
\begin{center}
\includegraphics[width=\textwidth]{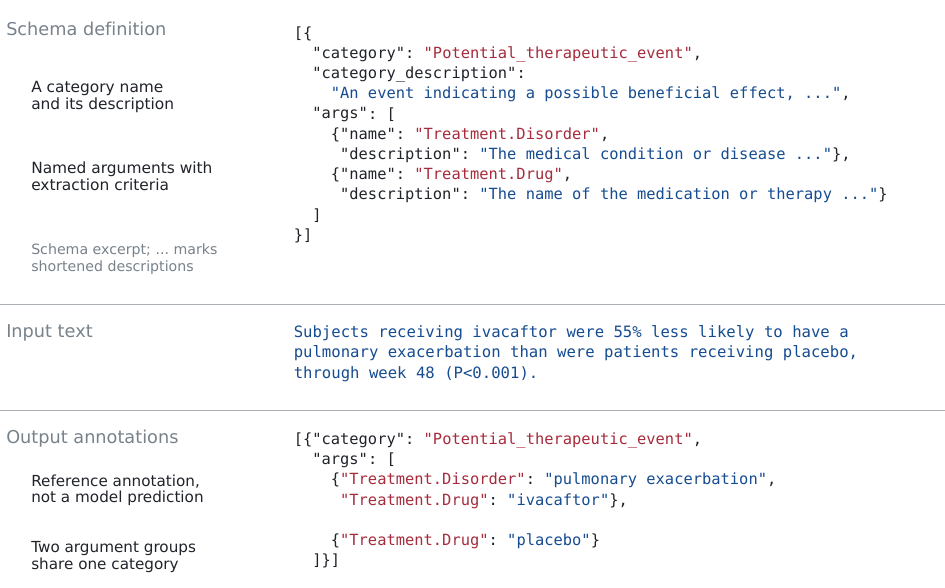}
\caption{An example of schema-conditioned input and structured annotation under the USP contract.}
\label{fig:dataconstruct}
\end{center}
\end{figure*}

\section{\ours}
\label{sec:method}
USP defines a shared semantic contract for corpus normalization, schema-conditioned parsing, and evaluation. Category labels paired with groups of named argument values provide a common output representation while retaining instance associations supplied by the source annotations.

Figure~\ref{fig:dataconstruct} illustrates the contract through a schema, input text, and reference annotation. Two argument groups remain separate under one category, preserving which treatment is associated with the annotated disorder rather than reducing the output to unbound triples. Construction details are provided in Appendix~\ref{apd:iepile}.

\subsection{Problem Formulation}
\label{sec:formulation}
Given an input utterance or dialogue $x$ and a task-specific schema $\mathcal{S}$, schema-driven semantic parsing produces an unordered collection of instantiated frames $y=\mathcal{F}(x,\mathcal{S})$, retaining repeated instances. NER identifies typed mentions; RE associates typed entity pairs; EE groups role arguments and, when supplied, triggers into event instances; and intent-slot parsing constructs intent or tool frames with slot values. Source datasets may provide intent-only, joint intent-slot, or dialogue-state annotations; their task designation follows the annotations used in the evaluation.

\subsection{The USP Semantic Contract}
\label{sec:contract}
A task schema contains $M$ category definitions,
\begin{equation}
\begin{aligned}
\mathcal{S}&=\{s_i\}_{i=1}^{M},\qquad
s_i=(c_i,d_i,\mathcal{A}_i),\\
\mathcal{A}_i&=\{(a_{ij},d_{ij})\}_{j=1}^{m_i},
\end{aligned}
\label{eq:usf-schema}
\end{equation}
where $c_i$ and $d_i$ are a category name and description, while $a_{ij}$ is an argument name and $d_{ij}$ is an optional extraction criterion when the source schema provides one. An instantiated frame is
\begin{equation}
\begin{aligned}
f_{i\ell}&=(c_i,\ell,\mathcal{V}_{i\ell}),\\
\mathcal{V}_{i\ell}&=\big\{\!\big\{(a_{ij},v_{i\ell jk})\big\}\!\big\}_{j,k},
\end{aligned}
\label{eq:usf-frame}
\end{equation}
where $\ell$ denotes the conceptual identity of a frame or call instance, not a required serialized identifier. Each output category object contains a list of argument groups, with one group per instance. Values may be extracted from the input or, where the task provides the necessary context and conventions, normalized or resolved from dialogue history. NER represents each entity through a \texttt{mention} field; RE preserves argument pairing; EE preserves instance boundaries and any source-provided trigger annotation; and intent-slot data preserve call or dialogue-state associations when available.

The USP representation serves as a semantic contract because the same schema definition governs corpus normalization, model conditioning, output construction, and evaluation. Its outputs are machine-consumable structured records. Parameter types, required-field constraints, enumerations, and side effects remain application-specific, so the representation does not itself define an executable API or a new ontology.

\subsection{Operational Requirements of the Contract}
\label{sec:functional-view}
Parsing under the contract requires selecting the categories expressed in the input, grounding values for schema-defined fields, binding those values to the appropriate role and instance, and serializing the resulting frames. Grounding may involve extraction, normalization, or contextual resolution. These requirements describe one jointly generated output, not separately parameterized components or sequentially observed reasoning stages. All task families share the learning objective in \S\ref{sec:learning}.

\subsection{Corpus Normalization and Quality Control}
\label{sec:collection}
We normalize \textbf{61} public datasets: \textbf{29} NER, \textbf{16} RE, \textbf{9} EE, and \textbf{7} intent-slot sources. The corpus spans Chinese and English and includes multilingual and multi-turn sources. Preprocessing utilities perform deduplication, instruction assembly, and annotation-consistency repair. Appendix~\ref{apd:data-collection-clean} describes these operations and distinguishes them from proposed cleaning procedures.

We draw the evaluation set from the complete dataset collection and use a fixed, protocol-consistent test subset covering all four families for the main comparison. Every compared model has observations on this subset. The complete dataset inventory preserves partial coverage and missing values for audit. Appendix~\ref{apd:data-collection-clean} defines each scope and Appendix~\ref{apd:preprocessing} describes the retained preprocessing components.

\subsection{Schema Descriptions}
\label{sec:augment}
Descriptions supply natural-language context for schema labels and optional argument criteria. Descriptions cover every category and most argument slots in the available prompts, but their specificity varies by family and is near-constant for NER. Coverage alone does not establish an accuracy benefit. Appendix~\ref{apd:description-augmentation} reports description coverage; controlled sensitivity to description content remains future work.

\subsection{Unified Learning Objective}
\label{sec:learning}
Given the serialized instruction $z=[I;\mathcal{S};x]$ and target tokens $y_{1:T}$, all task families share the autoregressive objective
\begin{equation}
\begin{aligned}
\mathcal{L}_{\mathrm{USP}}(\theta)={}&
-\sum_{(z,y)\in\mathcal{D}}\sum_{t=1}^{T}\\[-2pt]
&\log p_{\theta}(y_t\mid z,y_{<t}).
\end{aligned}
\label{eq:usf-loss}
\end{equation}
The task-specific schema conditions a shared token-prediction objective; the representation does not introduce a separate loss for each operational requirement.

\subsection{Structural Diagnostic Projections}
\label{sec:stage-eval}
The diagnostic interface examines the same prediction at different structural scopes. Category matching asks which categories are present; field matching checks category--argument pairs; field-value matching adds their values; instance-bound matching checks their association within an instance; and complete-instance matching requires the full instance to agree. Parse success and strict Equality Accuracy are reported separately. Each view has its own projection and counting unit, so the scores are neither processing-stage pass rates nor guaranteed to be mathematically nested. Task-native metrics remain necessary for direct comparison with prior systems.

For contract-level diagnosis, a parsed output is projected to
\begin{equation}
\begin{aligned}
\Pi(y)={}&\biguplus_{f_{i\ell}\in y}R(f_{i\ell}),\\[-2pt]
R(f_{i\ell})={}&
\big\{\!\big\{(\ell,c_i,a,v):
(a,v)\in\mathcal{V}_{i\ell}\big\}\!\big\},
\end{aligned}
\label{eq:field-projection}
\end{equation}
where $\ell$ is retained whenever the source annotation identifies a relation, event, state, or call instance. This expression specifies the instance-preserving record representation. The individual diagnostic views apply their own projection and matching rules, defined in Table~\ref{tab:metric-definitions-main-ie} in Appendix~\ref{sec:appendix-main-ie-full}. Dataset-specific task-standard scores may instead use their established projection; all reported tables must state which metric family they use.

For a category with no extracted fields, the field-level evaluator creates $(c_i,\texttt{NO\_SLOT},\texttt{NO\_VALUE})$ as a matching atom. An explicitly named field with an empty value is standardized to \texttt{NO\_VALUE}; omitted fields are not synthesized. These evaluation conventions are distinct from the model's serialized target. Task-standard metrics and USP diagnostic projections are complementary rather than interchangeable.

\subsection{Data Statistics}
\label{sec:stats}
The corpus inventory spans four task families, multiple domains, and Chinese and English sources. The final corpus retains about 2.75M instruction--output instances combining the 61 public datasets and the in-house XingChen annotations; task-level splits are 0.82M (NER), 0.91M (RE), 0.06M (EE), and 0.97M (intent--slot SLU). Per-task and per-dataset statistics are reported in Appendix~\ref{apd:iepile}; exact retained instance and token totals are given in Appendix~\ref{apd:iepile-stats} and on the released dataset page.


\section{Experiments}
\label{Experiments}
The evaluation addresses three questions. \textbf{RQ1: Contract conformance.} How consistently do converted annotations satisfy the contract, and where do violations concentrate? \textbf{RQ2: Comparative performance.} How do six models perform across the four task families under a shared protocol? \textbf{RQ3: Structural diagnosis.} What do the matching views and parse success reveal beyond the primary Field-Value score? Additional settings and unified-parsing results are in Appendix~\ref{apd:experiments}.

\subsection{Experimental Settings}
\textbf{Evaluation metrics.} Each diagnostic view $v$ defines its own matching units and returns counts $\mathrm{TP}_v$, $\mathrm{FP}_v$, and $\mathrm{FN}_v$. Category, field, and field-value views use set matching; instance-bound matching maximizes field overlap under one-to-one instance assignment; complete-instance matching uses multisets and retains multiplicity. These counts enter the same precision--recall calculation,
\begin{equation}
\begin{aligned}
P_v&=\frac{\mathrm{TP}_v}{\mathrm{TP}_v+\mathrm{FP}_v},\qquad
R_v=\frac{\mathrm{TP}_v}{\mathrm{TP}_v+\mathrm{FN}_v},\\
F_{1,v}&=\frac{2P_vR_v}{P_v+R_v}.
\end{aligned}
\label{eq:field-prf}
\end{equation}
with zero-denominator scores set to zero and results displayed as percentages. For NER, entity values are stored in the \texttt{mention} field; intent-slot projections use (\textit{intent}, \textit{slot-name}, \textit{slot-value}) records. Instance-aware views retain the grouping needed by relation, event, and multi-call outputs. Dataset-level micro scores pool matching counts over records with valid gold annotations before applying Eq.~\ref{eq:field-prf}. The frozen scorer treated both prediction-parsing failures and request failures as empty predictions for these counts. Its recorded parse-success measure also accepts empty responses, so it is not a request-success rate. We additionally report strict Equality Accuracy,
\begin{equation}
\begin{aligned}
\mathrm{EqAcc}=\frac{1}{N}\sum_{n=1}^{N}
\mathbf{1}\!\left[p_n\land C(\hat{y}_n)=C(y_n)\right],
\end{aligned}
\label{eq:eqacc}
\end{equation}
where $N$ counts records with valid gold annotations, $p_n$ indicates successful prediction parsing, and $C$ canonicalizes the complete instance multiset. Instance order is ignored, but repeated instances are retained; a parse failure cannot match a legal empty output. Table~\ref{tab:metric-definitions-main-ie} gives the view-specific definitions.
\textbf{Models.} The main comparison contains six models: five external references and \releasedmodel{}, fine-tuned from \backbonemodel{} using {\ours}. A4B denotes a mixture-of-experts model with 29B total parameters and 4B parameters activated per token; this describes active computation rather than memory footprint. \backbonemodel{} continues the TeleChat family of bilingual pretraining and alignment~\citep{he2024telechat,wang2024telechat}, dense-model post-training and reasoning~\citep{wang2025technical}, and MoE training infrastructure~\citep{liu2025training}; these reports provide historical context for the family.

\textbf{Evaluation scope and aggregation.} The main comparison uses a fixed test subset. Dataset-level Micro-F1 is macro-averaged within each task, and the four task macros receive equal weight in the overall score. Calculations use full precision before display rounding. Failed requests contribute empty predictions to aggregate scores. For \releasedmodel{}, CASIE requests timed out and RAMS requests exceeded the service context budget; their per-dataset cells are marked as unavailable in the appendix tables, while the aggregates include their zero contributions.

\subsection{Contract Conformance and Repairability}
\label{sec:contract-coverage}
RQ1 examines the converted annotations rather than model predictions. The 61-dataset inventory establishes the construction scope; the audit measures conformance of the available evaluation records, with source-to-USP semantic fidelity treated as a separate validation question.

\paragraph{Contract conformance.} We analyze $478{,}928$ gold evaluation records across $57$ dataset labels, some of which correspond to multiple datasets in the $61$-dataset construction inventory or to subsets thereof. Of these records, $90.66\%$ satisfy the contract, $6.11\%$ are invalid and $3.23\%$ are ambiguous. Conformance varies sharply across datasets. Of 57 datasets, 39 are at or above $95\%$ valid, ten are at or below $25\%$, and only eight lie between those bands. Table~\ref{tab:contract-conformance} gives the task-family breakdown.

\paragraph{Where the contract does not fit.} The low-conformance group is not distributed evenly across families, and its members fail for identifiable and largely dataset-specific reasons rather than through a gradual degradation of the contract. \texttt{MASSIVE}, \texttt{RAMS}, and \texttt{Wiki-ZSL} are recorded as entirely ambiguous because of schema handling during audit input preparation; parsed directly from assembled prompts, all three resolve as valid schemas. Four other datasets---\texttt{DuEE}, \texttt{DuEE\_Fin}, \texttt{CCFlaw}, and \texttt{MultiWOZ}---are each dominated by one recurring violation class. Event extraction and intent-slot understanding consequently have lower aggregate conformance than named-entity recognition and relation extraction, but five of the seven intent-slot datasets remain between $85\%$ and $100\%$ valid.

\paragraph{Repair classes.} The audit separates identifier disagreements from cases requiring semantic judgement. Of the $29{,}267$ invalid records, $52.6\%$ carry a dataset-specific but deterministic mapping---overwhelmingly \texttt{MultiWOZ} slot names that repeat their domain prefix---and a further $25.1\%$ differ from the schema only by letter case, separator or whitespace. The remaining $22.2\%$ require human adjudication. Including records classified as ambiguous, the audit assigns $95.41\%$ of the $478{,}928$ records to valid or mechanically repairable classes and $4.59\%$ to human review. The $21{,}963$ human-review records are concentrated, with \texttt{MASSIVE} and \texttt{Wiki-ZSL} accounting for $62\%$ of them. Within the audit-classified invalid records, identifier mismatches form the largest repair class; this does not establish source-annotation fidelity.

\paragraph{Validation scope.} Conformance measures structural validity rather than preservation of source semantics; a lossy conversion could still satisfy the output contract. This analysis covers available evaluation splits, not the training corpus, and classifies potential repairs without applying them. Source-to-USP fidelity requires separate validation.

\begin{table}[t]
\centering
\scriptsize
\setlength{\tabcolsep}{2.4pt}
\begin{tabularx}{\columnwidth}{@{}Xrrrrr@{}}
\toprule
Family & Datasets & Records & Valid & Invalid & Ambig. \\
\midrule
NER & 28 & 252{,}983 & 95.80 & 4.20 & 0.00 \\
RE  & 13 & 179{,}615 & 96.36 & 0.33 & 3.31 \\
EE  &  9 &  17{,}738 & 58.63 & 31.24 & 10.13 \\
Intent-slot & 7 & 28{,}592 & 29.30 & 43.68 & 27.01 \\
\midrule
All & 57 & 478{,}928 & 90.66 & 6.11 & 3.23 \\
\bottomrule
\end{tabularx}

\vspace{0.6em}
\begin{tabularx}{\columnwidth}{@{}Xrr@{}}
\toprule
Outcome for non-valid records & Records & \% all \\
\midrule
Deterministic dataset mapping & 15{,}408 & 3.22 \\
Case, separator, or spacing variant & 7{,}356 & 1.54 \\
Requires human adjudication & 6{,}503 & 1.36 \\
Multiple schema entries match & 15{,}460 & 3.23 \\
\midrule
Valid or mechanically repairable & 456{,}965 & 95.41 \\
\bottomrule
\end{tabularx}
\caption{Contract conformance on the evaluation splits. Percentages in the upper block are computed within each family; the lower block reports shares of all records. Conformance measures structural validity, not source-to-USP semantic fidelity.}
\label{tab:contract-conformance}
\end{table}

\subsection{Unified Schema-Driven Parsing Results}
\label{sec:main-ie-results}

\begin{table*}[t]
\centering
\small
\setlength{\tabcolsep}{4.1pt}
\begin{tabular}{l rrrr c rrrr c}
\toprule
 & \multicolumn{5}{c}{Field-Value Micro-F1} & \multicolumn{4}{c}{Other matching views} & Parsing \\
\cmidrule(lr){2-6} \cmidrule(lr){7-10} \cmidrule(lr){11-11}
Model & EE & NER & RE & IR & \textbf{Overall} & Category & Field & Inst.-bound & Complete & Parse \\
\midrule
\rowcolor{gray!10} \texttt{GLM-4-9B-Chat-HF} & 8.19 & 7.29 & 1.81 & 75.08 & 23.09 & 37.61 & 27.42 & 22.92 & 20.42 & 45.21 \\
\rowcolor{gray!10} \texttt{Qwen3-8B-Base} & 17.17 & 46.30 & 15.40 & 38.94 & 29.45 & 55.02 & 46.56 & 26.88 & 19.58 & 93.34 \\
\rowcolor{gray!10} \texttt{Qwen3.7-Flash} & 20.90 & 55.62 & 20.36 & 46.65 & 35.88 & 62.15 & 53.28 & 31.90 & 23.56 & 98.72 \\
\rowcolor{gray!10} \texttt{DeepSeek-V4-Flash} & 23.75 & 55.21 & 23.84 & 54.33 & 39.28 & 64.72 & 57.51 & 35.10 & 24.76 & 98.91 \\
\rowcolor{gray!10} \texttt{DeepSeek-V4-Pro} & 23.35 & 59.35 & 24.02 & 58.38 & 41.28 & 66.38 & 58.56 & 40.10 & 30.15 & 99.98 \\
\midrule
\rowcolor{blue!8} \releasedmodel{} & 39.83 & 71.99 & 38.06 & 80.92 & \textbf{57.70} & 69.31 & 65.43 & 57.44 & 50.88 & 97.98 \\
\bottomrule
\end{tabular}
\caption{Main comparison on the fixed test subset.}
\label{tab:main-comparison-anchor45}
\end{table*}

\begin{figure*}[t]
  \centering
  \includegraphics[width=\textwidth]{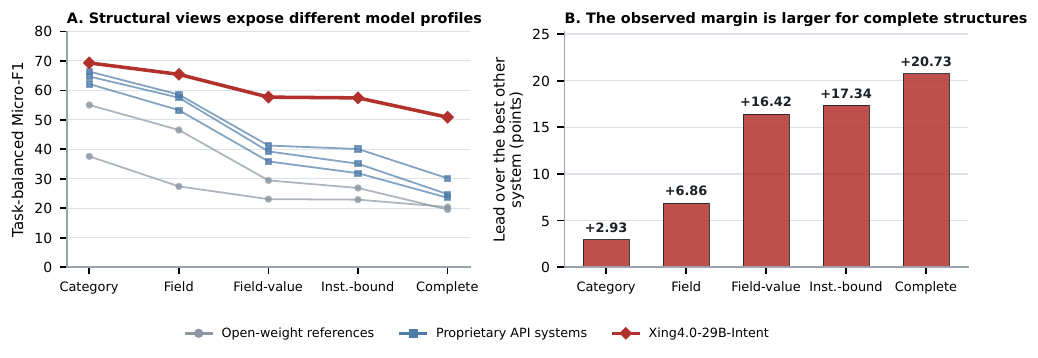}
  \caption{Task-balanced F1 under five structural matching views. Each view evaluates the same outputs using its own matching units; the profiles do not represent sequential processing stages.}
  \label{fig:stage-ladder-anchor45}
\end{figure*}

\begin{figure}[!t]
  \centering
  \includegraphics[width=\columnwidth]{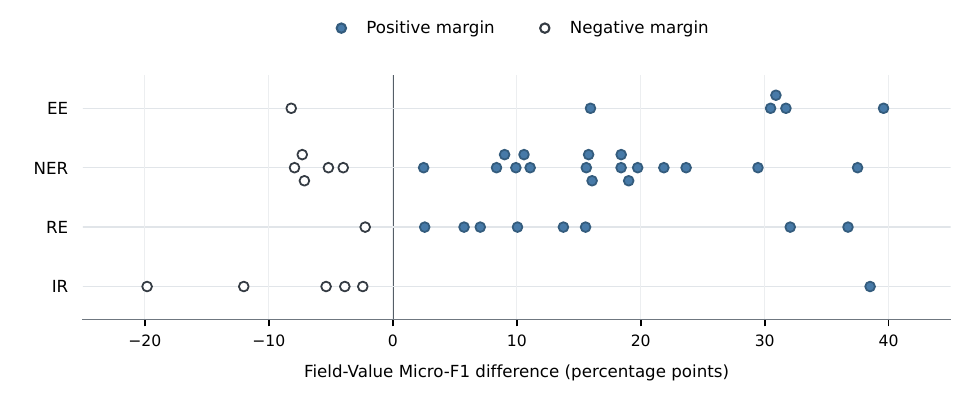}
  \caption{Per-dataset Field-Value Micro-F1 margins of \releasedmodel{} over the best alternative. Vertical offsets separate overlapping points within each task family.}
  \label{fig:dataset-margins-audited}
\end{figure}

\begin{figure*}[!t]
  \centering
  \includegraphics[width=\textwidth]{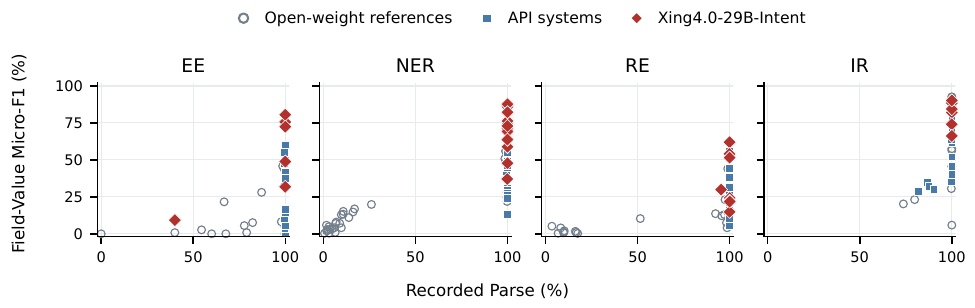}
  \caption{Parser acceptance and Field-Value Micro-F1 across model--dataset pairs, grouped by task. Parser acceptance includes empty outputs.}
  \label{fig:parse-field-quality-audited}
\end{figure*}

Table~\ref{tab:main-comparison-anchor45} compares six models under the evaluation protocol defined above. \releasedmodel{} has the highest score in every task family and F1 matching view, while \texttt{DeepSeek-V4-Pro} has the highest Parse rate.

The API results are restricted to the same fixed test subset as the other models. Family scores average dataset-level Field-Value Micro-F1, and Overall weights the four families equally.

The models show different task-family profiles. Section~\ref{sec:diagnostic-views} examines their dataset-level variation and agreement under different structural matching rules.

\subsubsection{Cost-Efficient Screening on USP-Core-6K}
\label{sec:core6k-screening}

USP-Core-6K extends model coverage at lower inference cost through outcome-blind, metadata-stratified sampling from the broader evaluation pool. It retains all four task families and every dataset identity in that pool. Selection uses task, language, input and schema length, gold-structure density, label frequency, and multiplicity, independently of model predictions or correctness. Table~\ref{tab:core6k-screening} reports 14 additional models under this reduced protocol, excluding those already shown in the main comparison. Supplementary runs with different service configurations are excluded from this ordering.

\begin{table}[t]
\centering
\scriptsize
\setlength{\tabcolsep}{1.5pt}
\begin{tabular*}{\columnwidth}{@{\extracolsep{\fill}}lrrrrrrr@{}}
\toprule
Model & EE & NER & RE & IR & \textbf{Overall} & Parse & Tokens \\
\midrule
\textrm{Gemini-3.6-Flash} & 30.46 & 70.73 & 29.34 & 64.34 & 48.72 & 99.40 & 22.34M \\
\textrm{Qwen3.8-Max} & 28.98 & 66.15 & 26.33 & 63.31 & 46.19 & 99.90 & 15.05M \\
\textrm{Kimi-K3} & 28.32 & 67.70 & 24.53 & 60.59 & 45.28 & 99.48 & 23.07M \\
\textrm{GPT-5.6-Sol} & 28.27 & 67.87 & 23.01 & 60.08 & 44.81 & 100.00 & 16.20M \\
\textrm{GLM-5.2} & 28.70 & 65.48 & 25.04 & 58.01 & 44.31 & 99.90 & 18.98M \\
\textrm{GPT-5.5} & 27.66 & 67.18 & 22.91 & 59.18 & 44.23 & 99.97 & 17.22M \\
\textrm{GLM-5.3-Flash} & 26.98 & 64.59 & 26.08 & 57.88 & 43.88 & 99.48 & 20.72M \\
\textrm{GLM-5.1} & 26.57 & 59.58 & 23.50 & 58.52 & 42.04 & 98.42 & 14.82M \\
\textrm{DeepSeek-V4.1-Flash} & 26.23 & 61.45 & 22.94 & 56.35 & 41.74 & 100.00 & 15.23M \\
\textrm{LongCat-2.0} & 24.23 & 62.29 & 22.79 & 53.36 & 40.67 & 99.23 & 19.99M \\
\textrm{GPT-4.1} & 23.11 & 58.73 & 21.56 & 55.18 & 39.65 & 99.90 & 15.41M \\
\textrm{GPT-6-Luna} & 24.24 & 59.19 & 20.00 & 50.93 & 38.59 & 99.52 & 15.38M \\
\textrm{GPT-4.1-Mini} & 21.03 & 56.54 & 21.46 & 48.71 & 36.94 & 99.82 & 15.34M \\
\textrm{GPT-4o-Mini} & 16.18 & 49.05 & 17.92 & 41.13 & 31.07 & 87.17 & 15.36M \\
\bottomrule
\end{tabular*}
\caption{Additional model results on USP-Core-6K, ordered by Overall.}
\label{tab:core6k-screening}
\end{table}

Among these additional models, \texttt{Gemini-3.6-Flash} has the highest observed score in each family and overall, followed by \texttt{Qwen3.8-Max} overall. The measurements broaden the comparison set; close rankings remain descriptive rather than significance results.

The screening and main comparisons use different test subsets, so their scores cannot be compared directly. Shared benchmark inputs and scoring do not imply equal inference budgets. Each external measurement reflects its provider configuration, so model, decoding, and service effects remain coupled rather than isolating model-family or generational effects.

Core-6K preserves the ordering of the three corresponding full-benchmark runs in the calibration set. We use it for initial screening, with close or anomalous cases requiring a larger suite. Appendix~\ref{sec:core6k-calibration} reports calibration errors and limitations.

\subsection{Structural Matching Analysis}
\label{sec:diagnostic-views}

A single Micro-F1 compresses disagreements about categories, values, and instance associations. Scoring the same outputs under category, field, field-value, instance-bound field, and complete-instance views makes these differences visible. Each view uses its own matching atoms and collection semantics; the scores answer distinct structural questions rather than forming a mathematically nested sequence or a new composite.

\paragraph{Value and instance agreement distinguish the profiles.} The SFT model leads under all five views, with a smaller margin for categories than for field values and complete instances. Figure~\ref{fig:stage-ladder-anchor45} shows that category agreement alone understates the differences in recovering values and complete structures.

\paragraph{Aggregate ordering masks local inversions.} The task-balanced aggregate descends across the views for all six models, but this pattern is empirical. Across the $1{,}080$ adjacent system--dataset--view-pair comparisons on the fixed test subset, the profile inverts $190$ times. The largest aggregate difference lies between field and field-value matching for five models. Four reference systems differ by $17.11$ to $18.22$ points between these views, compared with $7.73$ for the SFT row. This contrast locates a difference in value agreement under the evaluator, rather than an internal processing stage. Instance-bound matching adds a one-to-one assignment constraint and changes the counting units, so its F1 scores and those of field-value matching need not be ordered. The comparisons describe frozen rows without significance estimates. Appendix~\ref{sec:appendix-per-dataset} gives every test-subset cell.

\paragraph{Dataset-level variation.} Figure~\ref{fig:dataset-margins-audited} compares the SFT model with the highest-scoring alternative on each dataset, so the reference model may differ between points. Both positive and negative margins are shown. CASIE and RAMS are excluded because inference failed. The differences describe variation across datasets without estimating statistical significance.

\paragraph{Parsing and extraction quality.} Figure~\ref{fig:parse-field-quality-audited} relates Parse to Field-Value F1 for each model--dataset pair, excluding the SFT model's failed CASIE and RAMS runs. Models with similar Parse rates can have substantially different extraction scores. Because Parse also accepts empty outputs, these rates measure parser acceptance rather than successful inference.

\section{Discussion}
\label{sec:discussion}

\paragraph{What unification provides.}
USP supplies a common language for examining extraction and intent-slot outputs. Category labels identify the requested semantic type, named fields specify the relevant roles, and instance associations determine which values belong together. This common contract makes the same diagnostic questions available across families. Its value in the present study is representational and analytical; superiority to native task formulations would require a controlled comparison of both representations and evaluators.

\paragraph{Why one score is insufficient.}
The structural profile separates category agreement from recovery of values and instance associations. Differences between models depend on which unit is matched, making the profile more informative than one score. Execution status provides complementary information because parser acceptance alone does not establish that a response was received. Each matching view uses its own counting units, so differences between views describe structural agreement rather than pass rates through successive processing stages.

\paragraph{Where the four-family abstraction is strained.}
The common representation does not remove differences between annotation schemes. NER usually associates a type with a mention, whereas relations, events, and intent-slot records can bind several values within one instance. Description specificity and annotation conventions also vary across families. The conformance audit characterizes how converted records fit the contract; establishing that conversion preserves the source semantics requires a separate round-trip audit.

\paragraph{Robustness and external validity.}
Schema wording and input perturbations are separate dimensions of validity, as illustrated by SGD-X and RUIE-Bench~\citep{sgdx2022,zhu-etal-2025-towards}. The present fixed-schema comparisons do not measure those dimensions. They also use a USP-specific representation and evaluator, so published UIE, IEPile, KnowCoder, and task-specific scores are contextual references rather than directly comparable leaderboard entries. The untested controls are detailed in Limitations.

\paragraph{Supplementary general capability.}
MMLU-Pro results obtained with one fixed scoring protocol are reported in Appendix~\ref{apd:mmlu-track-i} as a separate general-capability reference. They neither explain the unified-parsing results nor supply a controlled base-versus-SFT contrast.

\section{Conclusion}
We presented {\ours}, a shared semantic contract for information extraction and intent-slot parsing. Its diagnostic views distinguish category agreement, field-value recovery, and instance associations across task families. The six-model comparison shows how model profiles change with the structural unit being evaluated. By pairing a common representation with explicit matching rules, USP supports more informative analysis than a single aggregate score.

\section*{Limitations}
\label{adx:limitation}
{\ours}'s value in this study is representational and analytical. This section details the controls that would strengthen causal claims but remain untested, together with the coverage boundaries of the corpus and evaluation protocol. First, conversion fidelity. The conformance audit characterizes whether converted records satisfy the output contract, not whether conversion preserves source semantics; a round-trip audit that converts USP records back to task-native formats and compares them with original annotations would be required. Second, schema-conditioning effects. The present comparisons use fixed schemas, so sensitivity to schema wording and description content is not measured; controlled perturbations of description text remain future work. Third, inference budgets are not equalized across models; the comparison therefore characterizes performance under the evaluated execution conditions rather than under a cost-matched protocol. Fourth, the contract does not remove differences between annotation schemes across families. NER associates a type with a single mention, whereas relation, event, and intent-slot records can bind multiple values within one instance; description specificity and annotation conventions also vary across families. Cross-family comparisons under USP therefore describe agreement with a shared contract rather than equivalence of task difficulty.
\bibliography{custom}

\appendix

\clearpage

\makeatletter
\setlength{\@fptop}{0pt}
\setlength{\@dblfptop}{0pt}
\setlength{\@fpsep}{12pt plus 2pt minus 2pt}
\setlength{\@dblfpsep}{12pt plus 2pt minus 2pt}
\makeatother

\section{Construction Details of {\ours}}
\label{apd:iepile}

\subsection{Data Collection and Cleaning}
\label{apd:data-collection-clean}

\paragraph{Data Collection.} To broadly cover domains and practical demands, we collect open datasets from multiple sources and re-format them into the unified schema-driven label-generation format (\S\ref{sec:formulation}).
{\ours} spans four tasks---NER, RE, EE, and Intent/Slot SLU---and includes bilingual (Chinese/English), multilingual (MASSIVE~\citep{massive2022}), and multi-turn dialogue (MultiWOZ~\citep{multiwoz2018}) sources.
The construction inventory contains \textbf{61} datasets (29 NER / 16 RE / 9 EE / 7 intent-slot). The main comparison uses a fixed test subset for all six models, with failed inference identified separately. Tables~\ref{statistic-ner}--\ref{statistic-ee} list the construction inventory and available citations.

\paragraph{Per-Task Dataset Inventory.}
The 61 datasets span NER, RE, EE, and Intent/Slot SLU across two primary languages and multiple domains. Tables~\ref{statistic-ner}--\ref{statistic-ee} list their source language, schema count, and available citation. NCBI-Disease, Wiki-ZSL, and CMeIE are outside the fixed test subset because of documented task or split fallback; the remaining starred rows are outside the test subset because of partial model coverage. MASSIVE is marked separately because it was normalized into the construction inventory but excluded from the current evaluation due to size.

\paragraph{Data Cleaning.}
The retained cleaning specification defines the following dataset-level operations, but the available records do not show that this complete sequence produced the reported build.
First, it calls for computing text overlap within a dataset's train/validation/test splits and removing texts that appear multiple times with inconsistent annotations.
It further specifies heuristic filters for low-quality data:
\textbf{1)} non-alphabetic characters comprising more than 80\% of the text;
\textbf{2)} text shorter than five characters without any labels;
\textbf{3)} an excessive proportion (\textgreater 80\%) of stopwords.

\subsubsection{Preprocessing Utilities}
\label{apd:preprocessing}

The retained preprocessing implementation contains three conceptual operations. \textbf{Structural deduplication} removes repeated field--value pairs, duplicated argument groups, and repeated top-level instances without changing unique annotations. \textbf{Instruction assembly} combines the source text, task-scoped schema, and structured target into one model-ready record while preserving task, language, domain, and split metadata. \textbf{Consistency repair} normalizes documented naming differences such as case, separators, repeated domain prefixes, and a small set of verified schema typos; ambiguous collisions are left unchanged for manual review.

The utilities implement these operations independently, and runtime inference and scoring are separate from preprocessing.

\paragraph{Validation coverage.}
The conformance analysis covers the available evaluation records, whose dataset labels are aggregated at the granularity each source provides; it is therefore a measure of the available records rather than a complete partition of the corpus.


\subsection{Schema Description Assets}
\label{apd:description-augmentation}
\label{apd:instruction-generation}

\paragraph{Schema Descriptions.}
Across 94,636 available prompts, all 1,098 category slots and 3,354 of 3,388 argument slots have non-empty descriptions. The 34 empty argument descriptions occur in intent-slot data. Description specificity varies substantially across families. In particular, the 182 NER argument slots use only two distinct descriptions, making coverage a poor proxy for category-specific information. These statistics describe prompt content; they do not measure the effect of descriptions on prediction quality.

\paragraph{Proposed Description-Sensitivity Study.}
A future study could hold the input, output contract, model, and prompt construction fixed while comparing name-only schemas, original descriptions, within-dataset scrambled descriptions, and length-matched filler. Reviewed paraphrases would provide an additional robustness check. This design tests inference-time sensitivity rather than the effect of descriptions during training.

\paragraph{Instruction Format.}
Each instruction is a structured record containing (1) the task description, (2) the queried categories and optional descriptions, and (3) the source utterance.
The target is a list of objects with a category label and an argument-group list. Each argument group maps schema-defined field names to their values and represents one instance. Unselected categories and unobserved fields are not filled with placeholder values. A selected category without extracted arguments uses an empty argument list. The evaluator introduces empty-slot markers internally when constructing diagnostic matching units; these are not required output fields. Table~\ref{tab:format} gives examples across all four task families.
Descriptions belong to the conditioning schema and are not repeated in the target annotation. Every argument displayed in the examples has a description; the excerpts show only the categories and fields needed to interpret the accompanying annotation, not the full candidate schema. This illustrative coverage does not imply that descriptions are present for every field in the corpus.

\paragraph{Illustrative Intent-Slot Example.}
The corpus representation separates schema definitions from data instances. Figure~\ref{fig:slu-example} shows an intent-slot example with a category description, a named argument and its description, an input utterance, and a reference annotation. The schema excerpt includes the field used by the annotation; other candidates and fields are not displayed.

\begin{figure*}[t]
\begin{lstlisting}
Schema (excerpt):
[{"category": "GetWeather",
  "category_description":
    "Inquiries about weather conditions for a specific location or time.",
  "args": [{"name": "state",
            "description": "The state for which weather information is requested"}]}]

Input:
Human: whats the weather in ga

Reference:
[{"category": "GetWeather", "args": [{"state": "ga"}]}]
\end{lstlisting}
\caption{An intent-slot schema excerpt, input utterance, and reference annotation.}
\label{fig:slu-example}
\end{figure*}

\subsection{Data Statistics}
\label{apd:iepile-stats}
The released corpus distribution (XingChen-AGI/Xing4.0-29B-Intent-oss on Hugging Face) reports 2,753,000 total samples reconstructed from the 61 public datasets and the in-house XingChen annotations, spanning 19 domains and 52 languages. Each retained instance contains an \texttt{instruction} and an \texttt{output} field formatted for instruction tuning. Per-split instance counts and source URLs are given on the dataset page.\footnote{\url{https://huggingface.co/datasets/XingChen-AGI/Xing4.0-29B-Intent-oss}}

\section{Experiments}
\label{apd:experiments}

\subsection{Experimental Settings}

\paragraph{Evaluation Metrics.}
The reported comparison uses USP diagnostics: category, field, field-value, instance-bound field, complete instance, parse success, and strict Equality Accuracy. For matching views that require field-level atoms, the evaluator represents a category with no fields as $(c,\texttt{NO\_SLOT},\texttt{NO\_VALUE})$ and converts an explicitly named empty field value to \texttt{NO\_VALUE}. An omitted field is not automatically inserted. Task-native span, structure, intent, and slot metrics are not mixed into the main ranking; paired native-versus-USP scoring remains required for direct comparison with task-specific systems. MMLU-Pro Track~I is a separate general-capability reference and is not aggregated with unified parsing.

The main-text names Field, Inst.-bound, and Complete correspond to the retained evaluator's Key, Argument-binding, and Instance-aware full-event views, respectively. These are display aliases, not additional metrics. The definitions accompanying the per-dataset results specify the set, assignment, or multiset semantics used by each view.

\paragraph{Baselines and Recipes.}
The six-model unified-parsing comparison is defined in \S\ref{sec:main-ie-results}.

\subsection{Training Configuration}
The checkpoint in the main comparison was fine-tuned on clusters of Ascend 910C accelerators (80\,GB each). Fine-tuning ran for 3 epochs with a batch size of 32 and a learning rate of $1\text{e}{-5}$ under a cosine schedule with linear warm-up. The training mixture draws 100K stratified instances per collected dataset and blends them with a proportional share of general-purpose instructions.

\subsection{Unified Schema-Driven Parsing: Full Results}
\label{sec:appendix-main-ie-full}

This appendix defines the metrics and reports per-dataset results for the fixed test subset. The construction inventory lists all 61 datasets; its scope is distinct from the model comparison.

\subsubsection{Metrics and Aggregate Summary}
Table~\ref{tab:metric-definitions-main-ie} defines each matching view and states which structural unit it scores. The values behind Figure~\ref{fig:stage-ladder-anchor45} appear in Table~\ref{tab:main-comparison-anchor45} and are not repeated here.

Table~\ref{tab:main-comparison-anchor45} orders models by Overall, which weights the four family macros equally. All matching-view F1 scores use this aggregation. Parse likewise averages dataset-level rates within each family and then across families; it is not a pooled record share. The three API results are projected from $94{,}636$-item full-benchmark runs onto the fixed test subset. Failure handling follows the main experimental settings.

\begin{table*}[t]
\centering
\scriptsize
\setlength{\tabcolsep}{3pt}
\begin{tabularx}{\textwidth}{@{}p{0.18\textwidth}p{0.23\textwidth}p{0.22\textwidth}X@{}}
\toprule
Measure & Matching unit & Matching rule & Reporting role \\
\midrule
Category & category & Set overlap & Category selection \\
Field & (category, field) & Set overlap & Field selection \\
Field-Value & (category, field, value) & Set overlap & Primary extraction measure \\
Instance-Bound Field & fields within matched instances & One-to-one assignment maximizing field overlap & Instance association \\
Complete Instance & complete instance & Multiset overlap & Complete structure recovery \\
Full-event & complete instance & Same implementation as Complete Instance & Compatibility name; no additional view \\
Legacy complete instance & complete instance & Set overlap & Appendix compatibility comparison \\
Equality Accuracy & complete output per sample & Exact multiset match & Sample-level exactness \\
MMLU-Pro Accuracy & multiple-choice answer & Exact answer match & Separate general-capability evaluation \\
\bottomrule
\end{tabularx}
\caption{Metric definitions and reporting roles. Structural matching views report Micro-F1 within each dataset; task and overall scores macro-average these dataset-level values. Accuracy measures are reported separately.}
\label{tab:metric-definitions-main-ie}
\end{table*}

\subsubsection{USP-Core-6K Calibration and Scope}
\label{sec:core6k-calibration}

USP-Core-6K contains 6,000 records and is nested within an 18,000-record confirmation suite; both retain the dataset identities in the broader evaluation pool. Selection uses a fixed seed and only outcome-blind metadata. Table~\ref{tab:core6k-calibration} compares the reduced-suite score with the corresponding full-benchmark score for the three calibration models.

For Table~\ref{tab:core6k-screening}, Parse counts parser-accepted predictions, including empty predictions, over the fixed $6{,}000$-record denominator. Failed requests remain as empty predictions. Tokens reports provider usage attached to the retained predictions, not total billed usage across retries. All displayed models use dedicated Core-6K runs and the v2 evaluator with four-family aggregation; later v3 rescoring is not mixed into these results. Rows are ordered by Overall. The three full-run projections used for calibration are reported separately in Table~\ref{tab:core6k-calibration}.

The four added measurements use the same frozen inputs and system prompt, with a $4{,}096$-token completion limit. \texttt{Qwen3.8-Max} and \texttt{DeepSeek-V4.1-Flash} were requested with thinking disabled and temperature zero. \texttt{GLM-5.3-Flash} cannot disable thinking and uses the lowest available reasoning setting at temperature zero. \texttt{GPT-6-Luna} requests no reasoning but requires the service-default temperature of one, with the completion limit sent through its supported API parameter. Historical rows also differ in decoding and service settings, so the table does not control inference budgets across models.


\begin{table}[t]
\centering
\footnotesize
\setlength{\tabcolsep}{2.4pt}
\begin{tabularx}{\columnwidth}{@{}Xrrrr@{}}
\toprule
Model & Core & Full & $|\Delta|$ & MAE \\
\midrule
Qwen3.7-Flash & 36.37 & 35.94 & 0.43 & 2.25 \\
DeepSeek-V4-Flash & 38.84 & 39.14 & 0.30 & 2.63 \\
DeepSeek-V4-Pro & 40.83 & 41.20 & 0.37 & 2.28 \\
\bottomrule
\end{tabularx}
\caption{Calibration of the reduced suite against full-benchmark results from the same broader evaluation pool. $|\Delta|$ is the absolute difference in task-balanced F1, and MAE is the mean absolute per-dataset difference. The preserved ordering is evidence for screening utility, not universal validity.}
\label{tab:core6k-calibration}
\end{table}

The same four-family aggregation is used for both reduced and full runs. Core-6K preserves the ordering of the three calibration models, and its absolute task-balanced error remains below half a point for each. This does not establish universal reduced-suite validity: close or anomalous cases must be escalated to the confirmation suite or full benchmark. Core-6K results are therefore reported separately from the main-test ranking.

\subsubsection{Per-Dataset Results for the Main Comparison}
\label{sec:appendix-per-dataset}
Tables~\ref{tab:per-dataset-ee-anchor45}--\ref{tab:per-dataset-ir-anchor45} report per-dataset Field-Value Micro-F1, with column macros at the foot of each table. Dashes denote failed inference for \releasedmodel{} on CASIE and RAMS. As specified in the main experimental settings, these cases contribute zeros to aggregate scores; other numeric zeros denote scored predictions.

\begin{table*}[t]
\centering
\small
\setlength{\tabcolsep}{4.0pt}
\begin{tabular}{lrrrrrr}
\toprule
Event extraction & \releasedmodelheader{} & \texttt{DS-V4-P} & \texttt{DS-V4-F} & \texttt{Qwen3.7-F} & \texttt{Qwen3-8B} & \texttt{GLM-4-9B} \\
\midrule
CASIE & --- & 9.77 & \textbf{11.94} & 8.68 & 8.16 & 5.56 \\
CrudeOil & \textbf{48.90} & 8.70 & 6.57 & 9.31 & 4.82 & 2.69 \\
DuEE & \textbf{75.70} & 57.74 & 59.75 & 54.91 & 44.34 & 21.64 \\
DuEE-Fin & \textbf{80.55} & 47.00 & 49.64 & 47.46 & 45.82 & 28.06 \\
FewFC & \textbf{31.77} & 0.03 & 0.00 & 0.05 & 0.05 & 0.00 \\
PHEE & \textbf{72.45} & 41.97 & 41.56 & 37.61 & 32.52 & 7.58 \\
RAMS & --- & \textbf{4.16} & 3.84 & 3.72 & 0.84 & 0.00 \\
WikiEvt & 9.23 & \textbf{17.42} & 16.71 & 5.45 & 0.80 & 0.00 \\
\midrule
\emph{Dataset macro} & \textbf{39.83}$^\dagger$ & 23.35 & 23.75 & 20.90 & 17.17 & 8.19 \\
\bottomrule
\end{tabular}
\par\smallskip
\parbox{\textwidth}{\footnotesize --- indicates inference failure. $^\dagger$The macro includes failed requests as zero contributions.}
\caption{Field-Value Micro-F1 on the event extraction datasets in the fixed test subset. Models are ordered by recorded overall Field-Value F1; bold marks row maxima among the displayed scores.}
\label{tab:per-dataset-ee-anchor45}
\end{table*}

\begin{table*}[t]
\centering
\small
\setlength{\tabcolsep}{4.0pt}
\begin{tabular}{lrrrrrr}
\toprule
Named-entity recognition & \releasedmodelheader{} & \texttt{DS-V4-P} & \texttt{DS-V4-F} & \texttt{Qwen3.7-F} & \texttt{Qwen3-8B} & \texttt{GLM-4-9B} \\
\midrule
AnatEM & \textbf{74.91} & 43.20 & 45.45 & 41.44 & 43.80 & 10.98 \\
BC2GM & \textbf{76.39} & 60.78 & 44.62 & 59.28 & 53.98 & 13.29 \\
BC5CDR & \textbf{82.35} & 73.98 & 68.24 & 67.33 & 50.16 & 2.97 \\
BioRED & \textbf{76.85} & 65.77 & 60.14 & 61.35 & 49.50 & 3.39 \\
Boson & \textbf{68.65} & 27.12 & 12.93 & 31.16 & 28.12 & 4.12 \\
CN-AI & 58.69 & \textbf{63.88} & 58.43 & 56.60 & 43.42 & 7.92 \\
CN-Lit & 58.96 & \textbf{66.09} & 66.07 & 60.06 & 50.82 & 7.12 \\
CN-Mus & 71.00 & \textbf{78.30} & 72.48 & 72.90 & 63.92 & 14.77 \\
CN-Pol & 72.68 & \textbf{76.68} & 70.30 & 69.40 & 55.89 & 19.88 \\
CN-Sci & 63.52 & \textbf{71.45} & 67.96 & 61.54 & 47.75 & 16.85 \\
CoNLL03 & \textbf{86.97} & 58.51 & 58.69 & 63.31 & 49.21 & 5.84 \\
EduNER & \textbf{47.78} & 28.05 & 26.78 & 29.36 & 24.62 & 4.09 \\
FabNER & \textbf{37.08} & 27.15 & 25.82 & 24.02 & 21.90 & 0.98 \\
Harvey & \textbf{76.33} & 56.58 & 55.12 & 52.14 & 49.78 & 15.20 \\
JNLPBA & \textbf{69.17} & 58.58 & 57.21 & 57.25 & 42.44 & 3.23 \\
MITMov & \textbf{85.96} & 67.53 & 67.42 & 64.96 & 53.89 & 0.33 \\
MITRest & \textbf{73.00} & 51.13 & 49.41 & 46.50 & 30.31 & 0.44 \\
MSRA & \textbf{87.77} & 61.60 & 59.72 & 68.74 & 56.27 & 4.80 \\
Onto & \textbf{82.12} & 65.35 & 59.84 & 66.04 & 48.81 & 2.15 \\
Resume & \textbf{87.61} & 71.81 & 64.78 & 36.87 & 55.52 & 7.02 \\
WikiGold & \textbf{82.28} & 73.26 & 68.30 & 72.12 & 55.45 & 12.92 \\
WNUT17 & \textbf{63.73} & 58.80 & 54.95 & 61.24 & 42.99 & 2.05 \\
\midrule
\emph{Dataset macro} & \textbf{71.99} & 59.35 & 55.21 & 55.62 & 46.30 & 7.29 \\
\bottomrule
\end{tabular}
\caption{Field-Value Micro-F1 on the named-entity recognition datasets in the fixed test subset. Models are ordered by overall Field-Value F1; bold marks row maxima.}
\label{tab:per-dataset-ner-anchor45}
\end{table*}

\begin{table*}[t]
\centering
\small
\setlength{\tabcolsep}{4.0pt}
\begin{tabular}{lrrrrrr}
\toprule
Relation extraction & \releasedmodelheader{} & \texttt{DS-V4-P} & \texttt{DS-V4-F} & \texttt{Qwen3.7-F} & \texttt{Qwen3-8B} & \texttt{GLM-4-9B} \\
\midrule
CoNLL04 & \textbf{61.92} & 56.17 & 53.38 & 44.48 & 44.08 & 5.15 \\
DocRED & \textbf{29.87} & 25.23 & 27.29 & 24.68 & 10.33 & 0.26 \\
FewRel & 14.86 & 17.06 & \textbf{17.09} & 15.97 & 13.01 & 0.90 \\
FinRED & \textbf{24.23} & 10.47 & 9.59 & 5.77 & 4.11 & 0.78 \\
GIDS & \textbf{61.95} & 29.22 & 29.89 & 26.88 & 23.07 & 3.99 \\
HacRED & \textbf{54.02} & 38.47 & 38.40 & 31.64 & 13.63 & 1.66 \\
KBP37 & \textbf{22.32} & 11.37 & 12.26 & 11.43 & 7.49 & 1.14 \\
SciERC & \textbf{21.83} & 14.77 & 11.83 & 11.73 & 12.20 & 2.02 \\
SemE-RE & \textbf{51.55} & 13.45 & 14.83 & 10.65 & 10.64 & 0.38 \\
\midrule
\emph{Dataset macro} & \textbf{38.06} & 24.02 & 23.84 & 20.36 & 15.40 & 1.81 \\
\bottomrule
\end{tabular}
\caption{Field-Value Micro-F1 on the relation extraction datasets in the fixed test subset. Models are ordered by overall Field-Value F1; bold marks row maxima.}
\label{tab:per-dataset-re-anchor45}
\end{table*}

\begin{table*}[t]
\centering
\small
\setlength{\tabcolsep}{4.0pt}
\begin{tabular}{lrrrrrr}
\toprule
Intent-slot understanding & \releasedmodelheader{} & \texttt{DS-V4-P} & \texttt{DS-V4-F} & \texttt{Qwen3.7-F} & \texttt{Qwen3-8B} & \texttt{GLM-4-9B} \\
\midrule
ATIS & 88.44 & 76.73 & 82.55 & 71.77 & 57.60 & \textbf{92.31} \\
CAIS & 82.10 & 71.01 & 68.66 & 61.76 & 56.49 & \textbf{87.48} \\
MulWOZ & \textbf{84.40} & 45.89 & 44.19 & 35.33 & 30.59 & 5.89 \\
SMP19 & 74.13 & 45.63 & 34.78 & 28.61 & 20.23 & \textbf{86.15} \\
SMP20 & 66.20 & 40.16 & 31.95 & 30.36 & 23.14 & \textbf{86.02} \\
SNIPS & 90.22 & 70.86 & 63.86 & 52.07 & 45.60 & \textbf{92.64} \\
\midrule
\emph{Dataset macro} & \textbf{80.92} & 58.38 & 54.33 & 46.65 & 38.94 & 75.08 \\
\bottomrule
\end{tabular}
\caption{Field-Value Micro-F1 on the intent-slot understanding datasets in the fixed test subset. Models are ordered by overall Field-Value F1; bold marks row maxima.}
\label{tab:per-dataset-ir-anchor45}
\end{table*}

Table~\ref{tab:observed-best-counts-anchor45} counts the datasets on which each model has the highest score under the aggregate failure-handling policy. \releasedmodel{} leads in 31 cells. The counts are descriptive; observed maxima do not establish statistical significance.

\begin{table*}[t]
\centering
\small
\setlength{\tabcolsep}{5.0pt}
\begin{tabular}{lrrrrr}
\toprule
Model & EE & NER & RE & IR & \textbf{Overall} \\
\midrule
\textbf{\releasedmodel{}} & 5 & 17 & 8 & 1 & \textbf{31} \\
\texttt{DeepSeek-V4-Pro} & 2 & 5 & 0 & 0 & \textbf{7} \\
\texttt{GLM-4-9B-Chat-HF} & 0 & 0 & 0 & 5 & \textbf{5} \\
\texttt{DeepSeek-V4-Flash} & 1 & 0 & 1 & 0 & \textbf{2} \\
\midrule
\multicolumn{6}{l}{\footnotesize No highest-observed cell: \texttt{Qwen3-8B-Base}, \texttt{Qwen3.7-Flash}} \\
\bottomrule
\end{tabular}
\caption{Fractional observed-best counts for Field-Value Micro-F1, ordered by total. Tied maxima share the count equally. Counts include the main protocol's failure handling and do not indicate statistical significance.}
\label{tab:observed-best-counts-anchor45}
\end{table*}

\subsection{Supplementary MMLU-Pro Evaluation}
\label{apd:mmlu-track-i}

\paragraph{Motivation.} Table~\ref{tab:mmlu-track-i} reports MMLU-Pro under a deterministic scorer and one zero-shot prompt shared across all models. This supplementary analysis uses the same fine-tuned checkpoint as the unified-parsing comparison, so the two experiments profile one trained model rather than two. The protocol also differs from public leaderboard evaluation.

\begin{table*}[t]
    \centering
    \setlength{\tabcolsep}{6pt}
    \small
    \resizebox{\textwidth}{!}{%
    \begin{tabular}{l|c|c|cc|cc}
    \toprule
        Model & Params & Track & Accuracy & 95\% CI & Fmt.~Compl. & Parse Rate \\
    \hline
        \multicolumn{7}{l}{\emph{External comparators}} \\
        GLM-4-9B-Chat-HF        & 9\,B & I & 0.3286 & [0.3202,\,0.3369] & 0.6990 & 0.7415 \\
        Qwen3-8B-Base           & 8\,B & I & 0.5998 & [0.5911,\,0.6084] & 0.9704 & 0.9503 \\
        \hline
        \multicolumn{7}{l}{\emph{Reported SFT checkpoint}} \\
        \releasedmodel{} & 29/4\,B & I & \textbf{0.6290}$^{\dagger}$ & [0.6204,\,0.6377] & 0.8974 & 0.9015 \\
    \bottomrule
    \end{tabular}%
    }
    \caption{MMLU-Pro accuracy on $12{,}032$ questions under a shared zero-shot boxed-answer protocol. Empty or unparseable answers count as incorrect; confidence intervals use paired question-level bootstrap. Format completion detects a boxed marker, while Parse Rate measures answer extraction. Results are not directly comparable with the public five-shot leaderboard.}
    \label{tab:mmlu-track-i}
\end{table*}

\paragraph{Test set.} MMLU-Pro~\citep{DBLP:journals/corr/abs-2406-01574} contains \(12{,}032\) items across 14 subjects, with up to ten options per item. Every row in Table~\ref{tab:mmlu-track-i} is scored on the same frozen test set.

\paragraph{Inference protocol.} Each model receives the same zero-shot instruction requesting a reasoning trace followed by a boxed answer letter. Native chat templates are applied per model. Greedy decoding is used with a maximum generation length of 4,096 tokens; no additional sampling or stopping parameters are imposed.

\paragraph{Scoring.} A deterministic scorer extracts the answer letter under ordered rules; the first matching rule is used.
\begin{itemize}
\itemsep 0.15em
\item \textbf{R1 (BOXED\_STRICT)}: last \verb|\boxed{X}| where \(X \in\) A--J (case insensitive) followed by end-of-payload or a non-A--J character.
\item \textbf{R2 (BOXED\_LOOSE)}: same after unwrapping complete presentation wrappers \verb|\text{}|, \verb|\mathbf{}|, \verb|\mathrm{}|, \verb|\operatorname{}| and stripping leading \verb|*|/\verb|(|/\verb|[| markup runs; boundary uses ASCII-only alpha rejection.
\item \textbf{R3 (CORRECT\_FINAL\_IS)}: last occurrence of ``(the )?(correct$|$final) answer is X''.
\item \textbf{R4 (ANSWER\_COLON)}: last occurrence of ``Answer[:$\colon$]~X''.
\item \textbf{R5 (BOXED\_OPTION\_TEXT)}: last \verb|\boxed{|\,\emph{option-text}\,\verb|}| whose normalized content matches exactly one option string (normalization: complete-wrapper unwrap, math-delimiter strip, whitespace collapse).
\item \textbf{R6 (UNPARSEABLE)}: no rule fires; the item is counted as \emph{wrong}.
\end{itemize}
The denominator for every reported percentage is \(12{,}032\); unparseable and empty responses count as wrong. Gold answers are consulted only after extraction. No non-boxed option-text fallback and no semantic similarity or LLM-judge rescue is used.

\paragraph{Cross-validation.} Two independent implementations of the extraction rules produced identical per-record answers, parse outcomes, applied rules, and correctness decisions on the shared test set. They also agreed on adversarial edge cases and protocol-derived unit tests.

\paragraph{Statistics.} All confidence intervals in Table~\ref{tab:mmlu-track-i} come from question-level paired bootstrap (\(B=10{,}000\), seed \(20260729\)) resampling the shared 12{,}032 question IDs with common indices per iteration; cross-model comparisons therefore preserve per-question pairing.
 
\paragraph{Limitations.} These are internal zero-shot scores and are not comparable to the public MMLU-Pro leaderboard, which uses a different prompt and evaluator. Responses that reach the generation limit before emitting a recognized final answer count as wrong, so the reported values are conservative for models that produce longer reasoning traces.


\begin{table*}[htbp]
    \centering
    \setlength{\tabcolsep}{3pt}
    \begin{tabular}{llccl}
    \toprule
    Task & Dataset & Lang. & \#Schemas & Cite \\
    \midrule
    \multirow{23}*{NER-en}
    & AnatEM & en & 1 & \citet{DBLP:journals/bioinformatics/PyysaloA14} \\
    & BC2GM & en & 1 & \citet{smith-etal-2008-bc2gm} \\
    & BC4CHEMD\textsuperscript{$\ast$} & en & 1 & \citet{krallinger-etal-2015-chemdner} \\
    & NCBI-Disease\textsuperscript{$\ast$} & en & 1 & \citet{DBLP:journals/jbi/DoganLL14} \\
    & BC5CDR & en & 2 & \citet{li-etal-2016-bc5cdr} \\
    & HarveyNER & en & 4 & \citet{DBLP:conf/naacl/ChenXZH22} \\
    & CoNLL2003 & en & 4 & \citet{tjong-kim-sang-de-meulder-2003-introduction} \\
    & MITRestaurant & en & 8 & \citet{DBLP:conf/icassp/LiuPCG13} \\
    & MITMovie & en & 12 & \citet{liu-etal-2013-query} \\
    & FabNER & en & 12 & \citet{DBLP:journals/jim/KumarS22} \\
    & MultiNERD\textsuperscript{$\ast$} & en & 16 & \citet{DBLP:conf/naacl/TedeschiN22} \\
    & Ontonotes & en & 18 & \citet{hovy-etal-2006-ontonotes} \\
    & FindVehicle\textsuperscript{$\ast$} & en & 21 & \citet{DBLP:journals/corr/abs-2304-10893} \\
    & CrossNER\_Politics & en & 9 & \citet{DBLP:conf/aaai/Liu0YDJCMF21} \\
    & CrossNER\_Literature & en & 12 & \citet{DBLP:conf/aaai/Liu0YDJCMF21} \\
    & CrossNER\_Music & en & 13 & \citet{DBLP:conf/aaai/Liu0YDJCMF21} \\
    & CrossNER\_AI & en & 14 & \citet{DBLP:conf/aaai/Liu0YDJCMF21} \\
    & CrossNER\_Science & en & 17 & \citet{DBLP:conf/aaai/Liu0YDJCMF21} \\
    & Few-NERD\textsuperscript{$\ast$} & en & 66 & \citet{DBLP:conf/acl/DingXCWHXZL20} \\
    & WNUT-2017 & en & 6 & \citet{derczynski-etal-2017-results} \\
    & WikiGold & en & 4 & \citet{balasuriya-etal-2009-named} \\
    & JNLPBA & en & 5 & \citet{collier-kim-2004-introduction} \\
    & BioRED & en & 6 & \citet{luo-etal-2022-biored} \\
    \midrule
    \multirow{6}*{NER-zh}
    & MSRA NER & zh & 3 & \citet{DBLP:conf/acl-sighan/Levow06} \\
    & Resume NER & zh & 8 & \citet{DBLP:conf/acl/ZhangY18} \\
    & CLUE NER & zh & 10 & \citet{DBLP:journals/corr/abs-2001-04351} \\
    & Weibo NER\textsuperscript{$\ast$} & zh & 4 & \citet{DBLP:conf/emnlp/PengD15} \\
    & Boson\textsuperscript{$\ddagger$} & zh & 6 & \citet{bosonnlp-2015-ner} \\
    & EduNER & zh & 16 & \citet{li-etal-2023-eduner} \\
    \bottomrule
    \end{tabular}
    \caption{Statistics of the 29 NER datasets in {\ours}: task subgroup, source language, per-dataset schema count from the dataset inventory, and the originating publication. NCBI-Disease is outside the fixed test subset because of documented task or split fallback; the other datasets marked \textsuperscript{$\ast$} are outside the test subset because of partial model coverage. Datasets marked \textsuperscript{$\ddagger$} do not have a verified origin publication. Per-split instance counts and source URLs are provided on the dataset page.}%
    \label{statistic-ner}
\end{table*}

\begin{table*}[htbp]
    \centering
    \setlength{\tabcolsep}{3pt}
    \begin{tabular}{llccl}
    \toprule
    Task & Dataset & Lang. & \#Schemas & Cite \\
    \midrule
    \multirow{10}*{RE-en}
    & ADE Corpus & en & 1 & \citet{gurulingappa-etal-2012-development} \\
    & GIDS & en & 4 & \citet{DBLP:conf/akbc/JatKT17} \\
    & CoNLL2004 & en & 5 & \citet{roth-yih-2004-linear} \\
    & SciERC & en & 7 & \citet{DBLP:conf/emnlp/LuanHOH18} \\
    & Semeval-RE & en & 10 & \citet{hendrickx-etal-2010-semeval} \\
    & KBP37 & en & 18 & \citet{DBLP:journals/corr/ZhangW15a} \\
    & Wiki-ZSL\textsuperscript{$\ast$} & en & 113 & \citet{DBLP:conf/naacl/ChenL21} \\
    & FewRel & en & 80 & \citet{DBLP:conf/emnlp/HanZYWYLS18} \\
    & FinRED & en & 29 & \citet{sharma-etal-2022-finred} \\
    & DocRED & en & 5/96 & \citet{yao-etal-2019-docred} \\
    \midrule
    \multirow{6}*{RE-zh}
    & CMeIE\textsuperscript{$\ast$} & zh & 53 & \citet{DBLP:conf/nlpcc/GuanZZXZ20} \\
    & DuIE2.0\textsuperscript{$\ast$} & zh & 49 & \citet{baidu-2020-duie2} \\
    & COAE2016\textsuperscript{$\ddagger$} & zh & 9 & \citet{coae-2016} \\
    & IPRE & zh & 35 & \citet{wang2019ipre} \\
    & SKE2020 & zh & 49 & \citet{baidu-2020-ske} \\
    & HacRED & zh & 26 & \citet{cheng-etal-2021-hacred} \\
    \bottomrule
    \end{tabular}
    \caption{Statistics of the 16 RE datasets in {\ours}: task subgroup, source language, per-dataset schema count from the dataset inventory, and the originating publication. Wiki-ZSL and CMeIE are outside the fixed test subset because of documented task or split fallback; DuIE2.0, also marked \textsuperscript{$\ast$}, is outside the test subset because of partial model coverage. Datasets marked \textsuperscript{$\ddagger$} do not have a verified origin publication and are cited by a shared-task placeholder. DuIE2.0 has no released training split. Per-split instance counts and source URLs are provided on the dataset page.}%
    \label{statistic-re}
\end{table*}

\begin{table*}[htbp]
    \centering
    \setlength{\tabcolsep}{3pt}
    \begin{tabular}{llccl}
    \toprule
    Task & Dataset & Lang. & \#Event types (\#Roles) & Cite \\
    \midrule
    \multirow{5}*{EE-en}
    & CASIE & en & 5 (26) & \citet{DBLP:conf/aaai/SatyapanichFF20} \\
    & PHEE & en & 2 (16) & \citet{DBLP:conf/emnlp/Sun0PWJGK022} \\
    & CrudeOilNews & en & 18 (104) & \citet{DBLP:conf/lrec/LeeSSS22} \\
    & RAMS & en & 106 (398) & \citet{ebner-etal-2020-multi} \\
    & WikiEvents & en & 31 (81) & \citet{DBLP:conf/naacl/LiJH21} \\
    \midrule
    \multirow{4}*{EE-zh}
    & DuEE1.0 & zh & 65 (217) & \citet{DBLP:conf/nlpcc/LiLPCPWLZ20} \\
    & DuEE-Fin & zh & 13 (91) & \citet{DBLP:conf/nlpcc/HanZLXPZ22} \\
    & FewFC & zh & 5 (29) & \citet{DBLP:conf/aaai/Zhou0ZWXL21} \\
    & CCF Law & zh & 9 (39) & \citet{ccf-bdci-2021-law} \\
    \midrule
    \multirow{7}*{IR / SLU}
    & MASSIVE\textsuperscript{$\dagger$} & multi & 60 & \citet{massive2022} \\
    & MultiWOZ2.2 & multi & 11 (61) & \citet{zang-etal-2020-multiwoz} \\
    & CAIS & zh & 14 (45) & \citet{liu2019cm} \\
    & SMP2019 & zh & 23 (108) & \citet{smp2019-ecdt} \\
    & SMP2020 & zh & 59 (153) & \citet{hou-etal-2022-fewjoint} \\
    & ATIS & en & --- & \citet{hemphill1990atis} \\
    & SNIPS & en & 7 & \citet{coucke2018snips} \\
    \bottomrule
    \end{tabular}
    \caption{Statistics of the 9 EE and 7 Intent/Slot SLU datasets in {\ours}: task subgroup, source language, per-dataset schema count from the dataset inventory (for EE, \#event types with \#argument roles in parentheses; the same convention applies to SLU intents), and the originating publication. MASSIVE, marked \textsuperscript{$\dagger$}, was normalized into the construction inventory but excluded from the current evaluation due to size. An unavailable schema count is shown as ``---'' rather than inferred; per-split instance counts and source URLs are provided on the dataset page.}%
    \label{statistic-ee}
\end{table*}

\begin{table*}[t]
\centering
\small
\setlength{\tabcolsep}{4pt}
\begin{tabular}{@{}p{0.07\textwidth}p{0.89\textwidth}@{}}
\toprule
\textbf{Task} & \textbf{Schema, input, and reference annotation} \\
\midrule
NER & \begin{lstlisting}[basicstyle=\ttfamily\footnotesize]
Schema (excerpt):
[{"category":"Person",
  "category_description":"People.",
  "args":[
    {"name":"mention","description":"Phrase or word segment corresponding to the entity"}
  ]}]
Input:
Coste said he had approached the player two months ago about a comeback .
Reference:
[{"category":"Person", "args":[{"mention":"Coste"}]}]
\end{lstlisting} \\
\midrule
RE & \begin{lstlisting}[basicstyle=\ttfamily\footnotesize]
Schema (excerpt):
[{"category":"Located in",
  "category_description":"Connects a smaller [Location] entity (like a city or facility) to a larger [Location] (like a state or country) that contains it.",
  "args":[
    {"name":"subject","description":"The source entity (Head) associated with the relation 'Located in'."},
    {"name":"object","description":"The target entity (Tail) associated with the relation 'Located in'."}
  ]}]
Input:
Locations containing suitable federally owned land were listed as : Fort Wainwright annex , Fairbanks , Alaska ;
Reference:
[{"category":"Located in", "args":[
  {"subject":"Fort Wainwright annex", "object":"Fairbanks"},
  {"subject":"Fort Wainwright annex", "object":"Alaska"},
  {"subject":"Fairbanks", "object":"Alaska"}]}]
\end{lstlisting} \\
\midrule
\bottomrule
\end{tabular}
\caption{USP schema and reference annotation examples across four task families. Schemas show relevant categories and fields; category and argument descriptions are reproduced in full for all displayed fields.}
\label{tab:format}
\end{table*}

\begin{table*}[t]
\ContinuedFloat
\centering
\small
\setlength{\tabcolsep}{4pt}
\begin{tabular}{@{}p{0.07\textwidth}p{0.89\textwidth}@{}}
\toprule
\textbf{Task} & \textbf{Schema, input, and reference annotation} \\
\midrule
EE & \begin{lstlisting}[basicstyle=\ttfamily\footnotesize]
Schema (excerpt):
[{"category":"movement-up-gain",
  "category_description":"An event describing an increase in price, value, or volume, without explicitly stating the cause.",
  "args":[
    {"name":"place","description":"The location of the movement."},
    {"name":"final_value","description":"The value after the upward movement."},
    {"name":"item","description":"The item affected by the upward movement."},
    {"name":"difference","description":"The amount of increase or gain."}
  ]}]
Input:
On the New York Mercantile Exchange , crude oil for delivery in November was up 1.36 % to $ 49.19 .
Reference:
[{"category":"movement-up-gain", "args":[
  {"place":"New York Mercantile Exchange", "final_value":"$ 49.19",
   "item":"crude oil", "difference":"1.36 %"}]}]
\end{lstlisting} \\
\midrule
Intent-slot & \begin{lstlisting}[basicstyle=\ttfamily\footnotesize]
Schema (excerpt):
[{"category":"AddToPlaylist",
  "category_description":"Requests to add a song, album, or artist to a specific playlist.",
  "args":[
    {"name":"artist","description":"The artist associated with the item to be added"},
    {"name":"playlist","description":"The name of the target playlist"}
  ]}]
Input:
add sabrina salerno to the grime instrumentals playlist
Reference:
[{"category":"AddToPlaylist", "args":[
  {"artist":"sabrina salerno", "playlist":"grime instrumentals"}]}]
\end{lstlisting} \\
\bottomrule
\end{tabular}
\caption{USP schema and reference annotation examples (continued).}
\end{table*}

\end{document}